\documentclass{article}

    \PassOptionsToPackage{numbers, compress}{natbib}
\usepackage{graphicx}
\usepackage{amsmath}
\usepackage{float}
\usepackage[utf8]{inputenc}
\usepackage{adjustbox}
 \usepackage[dblblindworkshop, final]{neurips_2025}
\workshoptitle{Evaluating the Evolving LLM Lifecycle: Benchmarks, Emergent Abilities, and Scaling}

\usepackage[utf8]{inputenc} % allow utf-8 input
\usepackage[T1]{fontenc}    % use 8-bit T1 fonts
\usepackage{hyperref}       % hyperlinks
\usepackage{url}            % simple URL typesetting
\usepackage{booktabs}       % professional-quality tables
\usepackage{amsfonts}       % blackboard math symbols
\usepackage{nicefrac}       % compact symbols for 1/2, etc.
\usepackage{microtype}      % microtypography
\usepackage{xcolor}         % colors
\title{SAGE: A Realistic Benchmark for Semantic Understanding}

\author{%
  Samarth Goel \\
  University of California, Berkeley\\
  Berkeley, CA 94704 \\
  \texttt{sgoel9@berkeley.edu} \\
  \And
  Reagan J. Lee \\
  University of California, Berkeley \\
  Berkeley, CA 94704 \\
  \texttt{reaganjlee@berkeley.edu} \\
  \AND
  Kannan Ramchandran \\
  University of California, Berkeley \\
  Berkeley, CA 94704  \\
  \texttt{kannanr@eecs.berkeley.edu} \\
}

\begin{document}

\maketitle

\begin{abstract}
As large language models (LLMs) achieve strong performance on traditional benchmarks, there is an urgent need for more challenging evaluation frameworks that probe deeper aspects of semantic understanding. We introduce SAGE (Semantic Alignment \& Generalization Evaluation), a rigorous benchmark designed to assess both embedding models and similarity metrics across five categories: Human Preference Alignment, Transformation Robustness, Information Sensitivity, Clustering Performance, and Retrieval Robustness. Unlike existing benchmarks that focus on isolated capabilities, SAGE evaluates semantic understanding through adversarial conditions, noisy transformations, and nuanced human judgment tasks across 30+ datasets. Our comprehensive evaluation of 9 embedding models and classical metrics reveals significant performance gaps, with no single approach excelling across all dimensions. For instance, while state-of-the-art embedding models like OpenAI's \verb|text-embedding-3-large| dominate in aligning with human preferences (0.682 vs. 0.591 for the best classical metric), they are significantly outperformed by classical metrics on information sensitivity tasks, where Jaccard Similarity achieves a score of 0.905 compared to the top embedding score of 0.794. SAGE further uncovers critical trade-offs: OpenAI's \verb|text-embedding-3-small| achieves the highest clustering performance (0.483) but demonstrates extreme brittleness with the lowest robustness score (0.011). SAGE exposes critical limitations in current semantic understanding capabilities and provides a more realistic assessment of model robustness for real-world deployment.\footnote{Code available: \url{https://github.com/sgoel97/neurips-2025-sage}}
\end{abstract}

\section{Introduction}

The rapid advancement of large language models has been accompanied by increasingly sophisticated benchmarks \cite{chang2023surveyevaluationlargelanguage}, yet current evaluation frameworks often fail to capture the nuanced, multifaceted nature of semantic understanding required for real-world applications. While benchmarks like MTEB \cite{muennighoff2023mteb} and BEIR \cite{thakur2021beir} provide valuable model rankings, they primarily assess performance under ideal conditions and focus narrowly on retrieval tasks, missing critical aspects of semantic robustness and human alignment. This gap becomes particularly problematic as AI systems are deployed in noisy, adversarial environments where robustness \cite{morris2020textattackframeworkadversarialattacks} and human alignment \cite{zhou-etal-2022-problems} are paramount \cite{magomere2025claimsevolveevaluatingenhancing}.

To address these limitations, we introduce SAGE (Semantic Alignment \& Generalization Evaluation), a benchmark designed around two core principles: Semantic Alignment (accuracy in reflecting human judgments and preferences) and Generalization (robustness under diverse and adversarial conditions). SAGE's unique contribution lies in its comprehensive evaluation of semantic understanding through deliberately challenging scenarios that expose model limitations invisible to traditional benchmarks.

\section{The SAGE Benchmark: A Holistic Evaluation Protocol}

SAGE is designed around two core principles required for deep semantic understanding:

\begin{enumerate}
    \item \textbf{Semantic Alignment:} The ability to accurately reflect human judgments, preferences, and the nuanced understanding of meaning.
    \item \textbf{Generalization:} Robustness and reliability when faced with diverse and challenging conditions, such as noisy data or adversarial perturbations.
\end{enumerate}

To measure these properties, SAGE aggregates performance across five distinct task categories, chosen to be more challenging and comprehensive than those in existing benchmarks.

\subsection{Task 1: Human Preference Alignment}

Semantic similarity metrics must reflect nuanced human judgments rather than surface-level patterns to be useful in real-world applications. This task ensures metrics align with how humans actually perceive and evaluate text quality and relevance.

We evaluate this alignment using OpenAI's human feedback dataset from the "summarize\_from\_feedback" collection \cite{stiennon2022learningsummarizehumanfeedback}. The dataset contains two types of human judgments: (1) multi-dimensional ratings of summary quality in the axis\_evals table, and (2) pairwise preferences between summaries in the comparisons table.

For multi-dimensional ratings, we measure how well a metric's similarity scores correlate with human quality assessments by computing the Pearson correlation between each summary-source similarity score and its corresponding human ratings. For pairwise preferences, we test whether the metric correctly predicts human choices - selecting the summary with higher similarity to the source as the preferred one - and measure prediction accuracy against ground truth preferences. Further details are provided in Appendix \ref{sec:task1}, and complete subtask scores are provided in \autoref{tab:sage-human-pref-comparisons} and \autoref{tab:sage-human-pref-scoring}.

\subsection{Task 2: Transformation Robustness}

Real-world text contains noise from OCR errors, typos, and formatting inconsistencies \cite{belinkov2018syntheticnaturalnoisebreak, pruthi2019combatingadversarialmisspellingsrobust}. A robust similarity metric should distinguish between superficial changes that preserve meaning and semantic alterations that fundamentally change content.

We evaluate this capability using three long-form text datasets: academic papers \cite{gupta-etal-2021-sumpubmed}, legislation \cite{kornilova-eidelman-2019-billsum}, and news articles \cite{see2017pointsummarizationpointergeneratornetworks}. For each document and its summary, we apply six transformations: three that preserve meaning (superficial perturbations like typos or synonym replacements) and three that alter meaning (semantic changes like negation or factual modifications).

We then measure three similarity relationships: original-to-superficial (after surface-level changes), original-to-semantic (after semantic changes), and original-to-summary (baseline summary similarity). A robust metric should consistently rank these relationships as: superficial perturbations maintain highest similarity, summaries have intermediate similarity, and semantic alterations show lowest similarity. We report the percentage of datapoints where all three relationships hold simultaneously. Implementation details and full task scores are in Appendix \ref{sec:task2} and \autoref{tab:sage-robustness}.

\subsection{Task 3: Information Sensitivity}

Similarity metrics should accurately detect and quantify semantic degradation \cite{goel2025quantifyingpositionalbiasestext, liu2020adversarialtraininglargeneural}. Unlike conventional robustness evaluations that test resilience to noise \cite{moradi2021evaluatingrobustnessneurallanguage}, this task measures whether metrics can precisely track how meaning changes as content is modified.

We test six long-form datasets by applying two perturbations: (1) inserting irrelevant content ("needle-in-haystack"), and (2) removing content spans. An information-sensitive metric should show similarity scores that decrease proportionally with the amount of perturbation - more inserted noise or removed content should yield correspondingly lower similarity scores.

We score each metric based on how closely its similarity changes follow this expected relationship: perfectly linear degradation receives the highest score, while erratic or flat responses score poorly. Detailed methodology and performance scores are in Appendix \ref{sec:task3} and both \autoref{tab:sage-sensitivity-insert} and \autoref{tab:sage-sensitivity-removal}.

\subsection{Task 4: Clustering Performance}

Effective similarity metrics should preserve meaningful categorical structure in unsupervised settings \citep{1427769}, making clustering a valuable proxy for semantic understanding. If a metric truly captures semantic relationships, documents with similar meanings should naturally cluster together.

We evaluate clustering quality across all 11 datasets from the Massive Text Embedding Benchmark (MTEB) \cite{muennighoff2023mteb}. Using agglomerative clustering with each model's similarity scores, we measure clustering quality via V-measure \cite{rosenberg-hirschberg-2007-v}. Complete details are provided in Appendix \ref{sec:task4}.

\subsection{Task 5: Retrieval Robustness}

Real-world retrieval systems must handle documents with various text corruptions including character-level noise, semantic alterations, and content contamination \cite{liu2024robustneuralinformationretrieval}. While traditional benchmarks assume clean corpora, practical applications require robustness to these common perturbations \cite{wu2022pradapracticalblackboxadversarial}.

We stress-test retrieval robustness using BEIR benchmark datasets \cite{thakur2021beir}. For each dataset, we create an adversarially augmented corpus by generating 18 perturbed versions of each document using transformations from tasks 2 and 3. These include 6 robustness transformations, 6 needle insertions at varying positions and sizes, and 6 content removals at varying positions and sizes.

Performance is measured as the ratio of NDCG@10 scores between each perturbed corpus and the original clean corpus. We report the harmonic mean of these retention ratios across all perturbations. Full implementation details are in Appendix \ref{sec:task5}.

\section{Experimental Setup}

We evaluated a suite of popular text embedding models and classical similarity metrics using the SAGE benchmark. Classical metrics included Levenshtein Ratio, ROUGE score \cite{lin-och-2004-automatic}, Jaccard similarity \cite{Jaccard1912}, and BM25 score. For all embedding models, we measured cosine similarity.

We used 5 SOTA embedding models based on various metrics such as industry adoption and performance on existing benchmarks. These include OpenAI's text-embedding-3-small and text-embedding-3-large, Cohere embed-v4.0, Voyage-3-large, and Gemini-embedding-001.

Scores for each task category are normalized to a 0-1 scale. The "Overall SAGE Score" is the unweighted average of the five category scores.

More information on each model and metric used can be found in appendix \ref{sec:models-metrics}.

\section{Results: Uncovering Nuanced Performance Trade-offs}

The results in \autoref{tab:sage-results-updated-retrieval} highlight SAGE's ability to reveal nuanced differences in performance that simpler benchmarks might miss. The central finding is that no single approach excels across all dimensions of semantic understanding.

\begin{table}
  \caption{Performance of embedding models and classical similarity metrics on the SAGE benchmark across five evaluation categories. Scores are normalized to [0,1]. The overall score is the unweighted mean across categories.}
  \label{tab:sage-results-updated-retrieval}
  \centering
  \begin{adjustbox}{center, max width=\textwidth}
  \begin{tabular}{lcccccc}
    \toprule
    Model / Metric & Clustering & Human Pref. & Robustness & Sensitivity & Retrieval & Overall \\
    \midrule
    \textbf{Embedding Models} & & & & & & \\
    embed-v4.0 & 0.396 & 0.648 & 0.070 & 0.789 & 0.389 & 0.458 \\
    gemini-embedding-001 & 0.387 & 0.674 & 0.319 & 0.725 & 0.417 & 0.504 \\
    text-embedding-3-large & 0.443 & \textbf{0.682} & 0.243 & 0.794 & \textbf{0.457} & \textbf{0.524} \\
    text-embedding-3-small & \textbf{0.483} & 0.654 & 0.011 & 0.794 & 0.426 & 0.474 \\
    voyage-3-large & 0.397 & 0.668 & 0.229 & 0.757 & 0.411 & 0.492 \\
    \midrule
    \textbf{Classical Metrics} & & & & & & \\
    BM25 Score & 0.209 & 0.591 & 0.283 & 0.673 & 0.240 & 0.399 \\
    Jaccard Similarity & 0.191 & 0.577 & 0.163 & \textbf{0.905} & 0.280 & 0.423 \\
    Levenshtein Ratio & 0.140 & 0.532 & \textbf{0.333} & 0.857 & 0.160 & 0.404 \\
    ROUGE Score & 0.190 & 0.568 & 0.178 & 0.875 & 0.200 & 0.402 \\
    \bottomrule
  \end{tabular}
  \end{adjustbox}
\end{table}

Among embedding models, OpenAI's text-embedding-3-large achieves the highest overall SAGE score (0.524), followed by gemini-embedding-001 (0.504) and voyage-3-large (0.492). Notably, all five embedding models substantially outperform classical similarity metrics, with the best-performing classical approach (Jaccard Similarity at 0.423) trailing the lowest-performing embedding model (text-embedding-3-small at 0.474) by a significant margin. 

Embedding models dominate in tasks requiring deep semantic understanding, such as human preference alignment (text-embedding-3-large: 0.682 vs. best classical BM25: 0.591), clustering (text-embedding-3-small: 0.483 vs. best classical BM25: 0.209), and retrieval (text-embedding-3-large: 0.457 vs. best classical Jaccard: 0.280). However, classical metrics demonstrate strong advantages in information sensitivity, where Jaccard Similarity achieves 0.905 compared to the top embedding score of 0.794, and in transformation robustness, where Levenshtein Ratio leads at 0.333 while embedding models top out at a score of 0.319. This trade-off reaches an extreme with text-embedding-3-small, which achieves the highest clustering performance (0.483) while simultaneously recording the lowest transformation robustness score across all approaches (0.011).

\section{Discussion: The Benchmark-Production Readiness Gap}

SAGE reveals a critical disconnect between benchmark performance and real-world readiness. While models achieve impressive scores on pristine datasets like MTEB and BEIR, our results show they fail under realistic conditions - text-embedding-3-small maintains only 1.1\% robustness despite strong clustering (0.483), and even the best retrieval model retains just 45.7\% effectiveness under adversarial noise. Real-world data is invariably corrupted through OCR errors, user typos, formatting inconsistencies, and transmission artifacts, yet current benchmarks evaluate only carefully curated inputs. This creates overconfidence: practitioners deploy models that excel on clean academic datasets but fail on the noisy production text that dominates real environments. Our finding that classical metrics outperform embeddings by 14\% on information sensitivity tasks directly contradicts MTEB rankings, demonstrating that benchmark leadership doesn't necessarily translate to success in all settings.

The performance variation across tasks underscores that model selection must account for both application requirements and data characteristics. Embedding models excel at clustering (2.3× better) and human preference alignment (15.4\% higher), yet classical metrics outperform them by 14\% on information sensitivity while robustness scores for embeddings can plummet to 0.011 under perturbation. These task-specific trade-offs explain many production failures: teams select models based on aggregate scores without understanding their brittleness. The 67\% failure rate of our most robust approach reveals that deploying embedding models without measures like domain-specific data cleaning, reranking, and filtering is premature for high-noise environments.

\section{Conclusion and Future Work}

In this work, we argued that a deeper form of semantic evaluation is needed to truly understand the capabilities of modern AI systems. We introduced the SAGE benchmark, a standardized protocol that assesses a wide range of technologies across a challenging and diverse set of tasks. Our results show that SAGE can uncover critical performance trade-offs, demonstrating that the optimal choice of model or metric is highly dependent on the specific application.

These findings demand a fundamental shift toward benchmarks that mirror production complexity. Future evaluations must incorporate real-world corruptions beyond our tested perturbations, with greater data diversity, adversarial augmentation by default, and production constraints like latency and memory limitations. Until such "production-strength" benchmarks exist, practitioners should assume published scores represent upper bounds achievable only in laboratory conditions and deploy accordingly with defensive architectures, ensemble methods, and appropriate safeguards for critical applications.

We hope SAGE will serve as a valuable tool for researchers and practitioners, fostering a more rigorous and balanced approach to evaluation.

%%%%%%%%%%%%%%%%%%%%%%%%%%%%%%%%%%%%%%%%%%%%%%%%%%%%%%%%%%%%

\bibliography{References}

%%%%%%%%%%%%%%%%%%%%%%%%%%%%%%%%%%%%%%%%%%%%%%%%%%%%%%%%%%%%

\appendix

\section{Technical Appendices and Supplementary Material}

\subsection{Dataset Details}
\label{sec:dataset_details}

\subsubsection{OpenAI Summarize from Feedback Dataset}
\label{sec:openai_summarize_feedback}

\paragraph{Source and Citation:} The Summarize from Feedback dataset was introduced by Stiennon et al. \cite{stiennon2022learningsummarizehumanfeedback} as part of their work on learning to summarize from human feedback. The dataset is publicly available through OpenAI and Hugging Face. The dataset includes machine-generated summaries for Reddit TL;DR posts, CNN articles, and Daily Mail articles. In total (across the comparisons and axis-eval parts) the Hugging Face release contains 193{,}841 rows.\,\,\cite{openai_summarize_hf} 

\paragraph{Structure and Size:} The dataset comprises two primary components used in our evaluation:
\begin{itemize}
    \item Axis Evaluations (\texttt{openai\_summarize\_scores}): Contains multi-dimensional human ratings of summary quality across various dimensions including overall quality, accuracy, coverage, and coherence. Summaries are rated on Likert scales.\,\,\cite{openai_summarize_hf,openai_summarize_repo}
    
    \item Pairwise Comparisons (\texttt{openai\_summarize\_comparisons}): Features 64{,}832 human preference judgments between summary pairs on the TL;DR dataset, where annotators indicate which of two summaries better represents the source text.\,\,\cite{openai_summarize_repo}
\end{itemize}

\subsubsection{Scientific Papers Dataset (PubMed)}
\label{sec:scientific_papers_pubmed}

\paragraph{Source and Citation:} The scientific papers dataset utilizes biomedical abstracts from the PubMed database, specifically leveraging the SumPubMed corpus introduced by Gupta et al. \cite{gupta-etal-2021-sumpubmed}. In our implementation, this appears as both \texttt{scientific\_papers\_sensitivity} and \texttt{scientific\_papers\_robustness} variants in \texttt{datasets.py}.

\paragraph{Structure and Size:} The dataset comprises 33{,}772 biomedical documents from the PubMed archive (BMC). Typical lengths in SumPubMed are: article \emph{raw-text} version averages 4{,}227 words / 203 sentences; corresponding summary averages 277 words / 14 sentences. For the noun-phrase and hybrid processed versions, the article averages are 1{,}578 and 1{,}891 words (with 57 and 71 sentences) and the summaries average 223 words (10 sentences).\,\,\cite{gupta-etal-2021-sumpubmed_pdf}

\subsubsection{CNN/DailyMail Dataset}
\label{sec:cnn_dailymail}

\paragraph{Source and Citation:} The CNN/DailyMail dataset was introduced by Hermann et al. and later refined by See et al. \cite{see2017pointsummarizationpointergeneratornetworks}. It is available through the \texttt{abisee/cnn\_dailymail} repository in our \texttt{datasets.py} configuration.

\paragraph{Structure and Size:} Contains over 300{,}000 unique news articles paired with human-written summaries:
\begin{itemize}
    \item News articles from CNN (Apr 2007–Apr 2015) and Daily Mail (Jun 2010–Apr 2015).\,\,\cite{cnn_dm_hf}
    \item Human-written summaries (“highlights”) for each article. Version 3.0.0 has splits: 287{,}113 train, 13{,}368 validation, 11{,}490 test; mean token counts: 781 per article and 56 per highlight.\,\,\cite{cnn_dm_hf}
\end{itemize}

\subsubsection{BillSum Dataset}
\label{sec:billsum}

\paragraph{Source and Citation:} The BillSum dataset was introduced by Kornilova and Eidelman \cite{kornilova-eidelman-2019-billsum} for summarization of U.S. Congressional and California state bills.

\paragraph{Structure and Size:} Contains 22{,}218 U.S. Congressional bills (103rd–115th Congress; 1993–2018), split into 18{,}949 train and 3{,}269 test, plus an additional California test set of 1{,}237 bills (2015–2016). Average preprocessed lengths: U.S. bills 1{,}382 words / 46 sentences (median 1{,}253 / 42), California bills 1{,}684 words / 47 sentences (median 1{,}498 / 42).\,\,\cite{billsum_pdf}

\subsubsection{Paul Graham Essays Dataset}
\label{sec:paul_graham_essays}

\paragraph{Source and Citation:} The Paul Graham Essays dataset consists of essays written by Paul Graham, available through the \texttt{sgoel9/paul\_graham\_essays} repository in our configuration.

\paragraph{Structure and Size:} The current Hugging Face release lists 215 rows (essays) in total.\,\,\cite{pg_hf}

\subsubsection{Amazon Polarity Dataset}
\label{sec:amazon_polarity}

\paragraph{Source and Citation:} The Amazon Polarity dataset is derived from Amazon product reviews, commonly used for sentiment classification tasks. It represents a subset of larger Amazon review datasets focusing on binary sentiment classification.

\paragraph{Structure and Size:} The original construction includes 3.6M training and 400k test reviews (two classes, balanced), using review \emph{title} and \emph{content} fields.\,\,\cite{zhang2015charcnn}

\subsubsection{ArguAna Dataset}
\label{sec:arguana}

\paragraph{Source and Citation:} The ArguAna dataset was introduced by Wachsmuth et al. \cite{wachsmuth-etal-2018-retrieval} and appears in the BEIR benchmark for argument retrieval.

\paragraph{Structure and Size:} In the BEIR format, ArguAna contains 8{,}674 documents and 1{,}406 queries; each query has on average 1.0 relevant document (Rel D/Q = 1.0).\,\,\cite{beir_wiki,beir_arguana_hf}

\subsubsection{MTEB Clustering Datasets}
\label{sec:mteb_clustering}

The Massive Text Embedding Benchmark (MTEB) clustering task comprises 11 datasets spanning diverse domains. These datasets provide ground-truth categorical labels for text segments across multiple domains. Several clustering tasks are provided in both sentence-to-sentence (S2S) and paragraph-to-paragraph (P2P) variants (titles only vs.\ title+content/abstract).\,\,\cite{mteb_paper}

Our evaluation utilizes the following datasets from the MTEB collection:

\paragraph{Scientific Literature Clustering:}
\begin{itemize}
    \item ArXiv Clustering (P2P and S2S): Scientific abstracts/titles from ArXiv. (S2S compares titles; P2P typically concatenates title+abstract.)\,\,\cite{mteb_paper}
    \item BioRxiv Clustering (P2P and S2S): Biomedical preprints. 
    \item MedRxiv Clustering (P2P and S2S): Medical preprints; the S2S/P2P datasets list 17{,}647 rows each in the Hugging Face release.\,\,\cite{mteb_medrxiv_s2s_hf,mteb_medrxiv_p2p_hf}
\end{itemize}

\paragraph{Community Discussion Clustering:}
\begin{itemize}
    \item StackExchange Clustering (Standard and P2P): Titles (standard) and title+body (P2P). The \emph{standard} variant comprises 25 sets, each with 10–50 classes and 100–1000 sentences per class; the \emph{P2P} variant provides 5 sets of 10k paragraphs and 5 sets of 5k paragraphs.\,\,\cite{mteb_stackex_standard_hf,mteb_stackex_p2p_hf}
    \item Reddit Clustering (Standard and P2P): Titles (standard) and title+posts (P2P). The \emph{standard} variant covers 199 subreddits across 25 sets (10–50 classes; 100–1000 sentences per class). The \emph{P2P} variant comprises 10 sets of 50k and 40 sets of 10k paragraphs.\,\,\cite{mteb_reddit_standard_hf,mteb_reddit_p2p_hf}
\end{itemize}

\paragraph{News and General Content Clustering:}
\begin{itemize}
    \item TwentyNewsgroups Clustering: Classical text classification dataset (\(\sim\) 18k posts across 20 topics). The MTEB clustering task uses subject-only text for clustering.\,\,\cite{twentynews_sklearn,mteb_twentyn_hf}
\end{itemize}

\subsubsection{BEIR Benchmark Datasets}
\label{sec:beir_benchmark}

\paragraph{Source and Citation:} The BEIR (Benchmarking Information Retrieval) benchmark was introduced by Thakur et al. \cite{thakur2021beir} as a heterogeneous benchmark for zero-shot evaluation of information retrieval models.

\paragraph{Structure and Size:} BEIR originally comprises 18 datasets (spanning 9 IR task types) standardized into \{corpus, queries, qrels\} format. Example scale points from the official stats: MS MARCO corpus has 8.84M documents and 6{,}980 dev/test queries; FEVER uses \(\sim\) 5.42M Wikipedia passages; TREC-COVID has 171k documents and 50 queries; ArguAna has 8.67k documents and 1{,}406 queries; CQADupStack has 457k documents and 13{,}145 queries.\,\,\cite{beir_wiki,beir_paper_table}

\subsection{Details for task 1: Human Preference Alignment}
\label{sec:task1}

\paragraph{Design Rationale:} Human preference alignment is central to evaluating semantic similarity. This task ensures that metrics reflect nuanced human judgments rather than surface-level similarity metrics, making them more suitable for real-world applications where human perception matters. By testing against both multi-dimensional quality ratings and pairwise preferences, we capture different aspects of how humans evaluate text similarity and quality.

\paragraph{Datasets:} We utilize OpenAI's human feedback dataset from the "summarize\_from\_feedback" collection \cite{stiennon2022learningsummarizehumanfeedback}, containing machine-generated summaries for Reddit TL;DR posts, CNN articles, and Daily Mail articles (see Section \ref{sec:openai_summarize_feedback} for full details). The dataset contains 193,841 total rows across all components \cite{openai_summarize_hf}. The axis\_evals table provides human ratings on a 1-7 Likert scale across four dimensions—overall quality, accuracy, coverage, and coherence—for summary-text pairs across the TL;DR and CNN/DM evaluation sets. The comparisons table contains 64,832 human preference judgments between summary pairs on the TL;DR dataset \cite{openai_summarize_repo}, where annotators selected their preferred summary or indicated a tie.

\paragraph{Evaluation:} We evaluate human preference alignment using two complementary approaches:

\textit{Pairwise comparison alignment (human\_pref\_comparisons):} Using the OpenAI summarize comparisons dataset, we predict human preferences by selecting the summary with higher similarity to the source text. For each similarity metric, we assign the preference to summary 1 if it has higher similarity than summary 2, otherwise to summary 2. Performance is evaluated using four classification metrics:
\begin{itemize}
\item Accuracy: Overall correctness of preference predictions
\item Precision: Proportion of correct predictions among all positive predictions
\item Recall: Proportion of correctly identified positive preferences
\item F1 Score: Harmonic mean of precision and recall
\end{itemize}

\textit{Multi-dimensional rating correlation (human\_pref\_scoring):} Using the OpenAI summarize scores dataset, we compute Pearson correlation between similarity scores (summary vs. source text) and human ratings across four quality dimensions:
\begin{itemize}
\item Overall: General quality assessment
\item Accuracy: Factual correctness of the summary
\item Coverage: Completeness of information capture
\item Coherence: Logical flow and readability
\end{itemize}

To normalize correlations to a [0, 1] scale suitable for comparison with other metrics, we apply the transformation:
\[
\text{score} = 0.5 \times (\text{correlation} + 1)
\]

This ensures all scores range from 0 (perfect negative correlation) to 1 (perfect positive correlation), with 0.5 representing no correlation.

\paragraph{Results} The results for each of the two datasets used in this task are presented below.

\begin{table}[H]
  \caption{Human Preference (pairwise comparisons) subtask metrics on SAGE.}
  \label{tab:sage-human-pref-comparisons}
  \centering
  \begin{adjustbox}{center, max width=\textwidth}
  \begin{tabular}{lcccc}
    \toprule
    Model / Metric & Accuracy & Precision & Recall & F1 \\
    \midrule
    \textbf{Embedding Models} & & & & \\
    embed-v4.0 & 0.668 & 0.668 & 0.670 & 0.669 \\
    gemini-embedding-001 & 0.702 & 0.707 & 0.691 & 0.699 \\
    text-embedding-3-large & \textbf{0.714} & \textbf{0.721} & \textbf{0.702} & \textbf{0.711} \\
    text-embedding-3-small & 0.668 & 0.674 & 0.654 & 0.664 \\
    voyage-3-large & 0.702 & 0.706 & 0.694 & 0.700 \\
    \midrule
    \textbf{Classical Metrics} & & & & \\
    BM25 Score & 0.574 & 0.577 & 0.569 & 0.573 \\
    Jaccard Similarity & 0.597 & 0.599 & 0.593 & 0.596 \\
    Levenshtein Ratio & 0.611 & 0.614 & 0.603 & 0.608 \\
    ROUGE Score & 0.580 & 0.581 & 0.582 & 0.582 \\
    \bottomrule
  \end{tabular}
  \end{adjustbox}
\end{table}

\begin{table}[H]
  \caption{Human Preference (scoring) subtask metrics on SAGE.}
  \label{tab:sage-human-pref-scoring}
  \centering
  \begin{adjustbox}{center, max width=\textwidth}
  \begin{tabular}{lcccc}
    \toprule
    Model / Metric & Overall & Accuracy & Coverage & Coherence \\
    \midrule
    \textbf{Embedding Models} & & & & \\
    embed-v4.0 & 0.662 & 0.599 & 0.662 & 0.587 \\
    gemini-embedding-001 & 0.692 & 0.620 & 0.688 & 0.593 \\
    text-embedding-3-large & \textbf{0.694} & \textbf{0.629} & \textbf{0.691} & \textbf{0.596} \\
    text-embedding-3-small & 0.685 & 0.613 & 0.683 & 0.591 \\
    voyage-3-large & 0.674 & 0.615 & 0.672 & 0.582 \\
    \midrule
    \textbf{Classical Metrics} & & & & \\
    BM25 Score & 0.654 & 0.577 & 0.662 & 0.544 \\
    Jaccard Similarity & 0.567 & 0.549 & 0.565 & 0.548 \\
    Levenshtein Ratio & 0.420 & 0.475 & 0.415 & 0.513 \\
    ROUGE Score & 0.562 & 0.547 & 0.558 & 0.551 \\
    \bottomrule
  \end{tabular}
  \end{adjustbox}
\end{table}

\subsection{Details for task 2: Transformation Robustness}
\label{sec:task2}

\paragraph{Design Rationale:} This evaluation framework specifically targets the brittleness observed in embedding models when confronted with character-level variations common in real-world text processing scenarios \cite{belinkov2018syntheticnaturalnoisebreak, pruthi2019combatingadversarialmisspellingsrobust}. Unlike conventional adversarial evaluations that primarily focus on semantic preservation \cite{ribeiro-etal-2020-beyond}, our transformation methodology maintains surface readability while testing whether models understand content semantically rather than relying on token-level patterns.

\paragraph{Datasets:} We utilize three long-form text corpora with distinct linguistic characteristics: biomedical abstracts from the SumPubMed corpus (Section \ref{sec:scientific_papers_pubmed}) containing 33,772 documents with average raw-text length of 4,227 words \cite{gupta-etal-2021-sumpubmed_pdf}, news articles from the CNN/DailyMail dataset (Section \ref{sec:cnn_dailymail}) using the 11,490 test split articles with mean length of 781 tokens \cite{cnn_dm_hf}, and legislative documents from the BillSum corpus (Section \ref{sec:billsum}) using the 3,269 test split U.S. bills with average length of 1,382 words \cite{billsum_pdf}. Together, these provide 48,531 document-summary pairs across formal academic, journalistic, and legal writing styles.

\paragraph{Evaluation:} We apply six systematically designed transformations:

\textit{Superficial perturbations (preserve meaning):}
\begin{itemize}
\item Random capitalization: 25\% of characters randomly capitalized (not case-switched)
\item Character deletion: Every 10th character removed (with special handling for spaces to preserve word boundaries)
\item Numerization: Character substitutions - 'e' → '3', 'i' → '1', 'a' → '4', 'o' → '0'
\end{itemize}

\textit{Semantic alterations (change meaning):}
\begin{itemize}
\item Negation toggling: Systematic reversal of affirmative/negative statements using regex patterns (e.g., ``is'' $\leftrightarrow$ ``is not'', ``can'' $\leftrightarrow$ ``cannot'')
\item Sentence shuffling: Random permutation of all sentences in the document
\item Word shuffling: Random permutation of all words within the entire text
\end{itemize}

For each document, we compute four similarity scores: original-to-original (baseline), original-to-superficial (averaged across three superficial perturbations), original-to-semantic (averaged across three semantic alterations), and original-to-summary. A robust metric should maintain the hierarchy: superficial similarity > summary similarity > semantic similarity. We evaluate three specific ordinal relationships:
\begin{itemize}
\item \texttt{summary\_over\_semantic}: Summary similarity exceeds all semantic alteration similarities
\item \texttt{superficial\_over\_summary}: All superficial perturbation similarities exceed summary similarity
\item \texttt{superficial\_over\_semantic}: All superficial perturbation similarities exceed all semantic alteration similarities
\end{itemize}
The final robustness score is the average percentage of test instances satisfying each of these three conditions.

\paragraph{Results} The results for each model, metric, and dataset are presented below.

\begin{table}[H]
  \caption{Robustness subtask metrics on SAGE.}
  \label{tab:sage-robustness}
  \centering
  \begin{adjustbox}{center, max width=\textwidth}
  \begin{tabular}{lccc}
    \toprule
    Model / Metric & BillSum & CNN/DailyMail & Scientific Papers
    \\
    \midrule
    \textbf{Embedding Models} & & & \\
    embed-v4.0 & 0.164 & 0.030 & 0.015 \\
    gemini-embedding-001 & 0.327 & \textbf{0.333} & 0.298 \\
    text-embedding-3-large & 0.311 & 0.295 & 0.123 \\
    text-embedding-3-small & 0.030 & 0.001 & 0.004 \\
    voyage-3-large & 0.316 & 0.324 & 0.047 \\
    \midrule
    \textbf{Classical Metrics} & & & \\
    BM25 Score & 0.256 & 0.260 & \textbf{0.333} \\
    Jaccard Similarity & 0.185 & 0.174 & 0.128 \\
    Levenshtein Ratio & \textbf{0.335} & \textbf{0.333} & 0.331 \\
    ROUGE Score & 0.225 & 0.145 & 0.163 \\
    \bottomrule
  \end{tabular}
  \end{adjustbox}
\end{table}

\subsection{Details for task 3: Information Sensitivity}
\label{sec:task3}

\paragraph{Design Rationale:} This evaluation task specifically measures semantic change detection - a critical capability for applications requiring fine-grained content monitoring such as document version-control systems, compliance tracking, and LLM watermarking \cite{yoo2023novel}. Unlike conventional robustness evaluations that assess resilience to noise \cite{moradi2021evaluatingrobustnessneurallanguage}, we evaluate whether similarity metrics can monotonically detect semantic noise or degradation and accurately reflect this in their output.

\paragraph{Datasets:} We use six diverse text domains with varying rhetorical structures: biomedical abstracts from PubMed (Section \ref{sec:scientific_papers_pubmed}) containing 33,772 documents averaging 4,227 words \cite{gupta-etal-2021-sumpubmed_pdf}, technological essays from the Paul Graham corpus (Section \ref{sec:paul_graham_essays}) with 215 essays \cite{pg_hf}, news articles from CNN/DailyMail test split (Section \ref{sec:cnn_dailymail}) with 11,490 articles averaging 781 tokens \cite{cnn_dm_hf}, consumer reviews from Amazon Polarity dataset (Section \ref{sec:amazon_polarity}) using 400,000 test reviews \cite{zhang2015charcnn}, legislative documents from BillSum test split (Section \ref{sec:billsum}) with 3,269 bills averaging 1,382 words \cite{billsum_pdf}, and argumentative texts from the ArguAna corpus (Section \ref{sec:arguana}) containing 8,674 documents \cite{beir_arguana_hf}. In total, we evaluate 457,420 documents across these domains.

\paragraph{Evaluation:} We apply two controlled perturbation strategies:

\textit{Irrelevant content insertion (``needle-in-haystack''):}
\begin{itemize}
\item Content source: Lorem Ipsum text of varying lengths from \url{https://www.lipsum.com/}
\item Proportions: 15\%, 50\%, 100\% of original document token length
\item Positions: Beginning (position 0), middle (position 0.5), end (position 1.0) of document
\item Implementation: Inserts needle text at exact character position calculated as \texttt{position * len(text)}
\end{itemize}

\textit{Token-based content removal:}
\begin{itemize}
\item Removal levels: 15\%, 50\%, 90\% of document tokens
\item Selection strategy: Contiguous token removal starting from position-adjusted locations
\item Position adjustment: Starting position is calculated as \texttt{position * (1 - removal\_size)} to account for document shortening
\item Implementation: Removes tokens at token-level using tokenizer encoding/decoding
\end{itemize}

We model expected similarity degradation using the theoretical relationship:

\[
\text{similarity} = 1 - \frac{p}{1+p},
\]

where \(p\) represents the perturbation proportion. This formula assumes diminishing marginal impact of additional perturbations, reflecting that initial changes have greater relative effect than subsequent ones.

\medskip

Performance is computed as:

\[
\text{sensitivity\_score} = 1 - \text{MAE},
\]

where MAE is the mean absolute error between observed and theoretical similarity values across all perturbation levels:

\[
\text{MAE} = \frac{1}{n}\sum_{i=1}^{n} \left| \text{observed}_i - \text{theoretical}_i \right|.
\]

Scores range from 0 (completely insensitive) to 1 (perfectly calibrated sensitivity).

\paragraph{Results} Results for each of the insertion and removal perturbations across models, metrics, and datasets are shown below.

\begin{table}[H]
  \caption{Sensitivity (insertion perturbation) subtask metrics on SAGE.}
  \label{tab:sage-sensitivity-insert}
  \centering
  \begin{adjustbox}{center, max width=\textwidth}
  \begin{tabular}{lcccccc}
    \toprule
    Model / Metric & Amazon Polarity & ArguAna & BillSum & CNN/DailyMail & Paul Graham & Scientific Papers \\
    \midrule
    \textbf{Embedding Models} & & & & & & \\
    embed-v4.0 & 0.749 & 0.716 & 0.701 & 0.700 & 0.706 & 0.710 \\
    gemini-embedding-001 & 0.698 & 0.699 & 0.719 & 0.702 & 0.702 & 0.698 \\
    text-embedding-3-large & 0.760 & 0.752 & 0.770 & 0.749 & 0.777 & 0.739 \\
    text-embedding-3-small & 0.809 & 0.782 & 0.765 & 0.739 & 0.751 & 0.753 \\
    voyage-3-large & 0.722 & 0.720 & 0.697 & 0.727 & 0.702 & 0.700 \\
    \midrule
    \textbf{Classical Metrics} & & & & & & \\
    BM25 Score & 0.596 & 0.604 & 0.642 & 0.755 & 0.734 & 0.685 \\
    Jaccard Similarity & \textbf{0.974} & \textbf{0.972} & \textbf{0.941} & \textbf{0.964} & \textbf{0.963} & \textbf{0.947} \\
    Levenshtein Ratio & 0.883 & 0.854 & 0.838 & 0.842 & 0.844 & 0.851 \\
    ROUGE Score & 0.888 & 0.884 & 0.881 & 0.881 & 0.881 & 0.882 \\
    \bottomrule
  \end{tabular}
  \end{adjustbox}
\end{table}

\begin{table}[H]
  \caption{Sensitivity (removal perturbation) subtask metrics on SAGE. }
  \label{tab:sage-sensitivity-removal}
  \centering
  \begin{adjustbox}{center, max width=\textwidth}
  \begin{tabular}{lcccccc}
    \toprule
    Model / Metric & Amazon Polarity & ArguAna & BillSum & CNN/DailyMail & Paul Graham & Scientific Papers \\
    \midrule
    \textbf{Embedding Models} & & & & & & \\
    embed-v4.0 & \textbf{0.894} & \textbf{0.877} & 0.844 & 0.846 & 0.864 & \textbf{0.868} \\
    gemini-embedding-001 & 0.765 & 0.750 & 0.738 & 0.740 & 0.747 & 0.738 \\
    text-embedding-3-large & 0.879 & 0.856 & 0.810 & 0.812 & 0.806 & 0.821 \\
    text-embedding-3-small & 0.878 & 0.852 & 0.783 & 0.802 & 0.794 & 0.821 \\
    voyage-3-large & 0.856 & 0.807 & 0.812 & 0.806 & 0.763 & 0.765 \\
    \midrule
    \textbf{Classical Metrics} & & & & & & \\
    BM25 Score & 0.591 & 0.595 & 0.639 & 0.779 & 0.763 & 0.693 \\
    Jaccard Similarity & 0.829 & 0.835 & \textbf{0.872} & 0.849 & 0.864 & 0.852 \\
    Levenshtein Ratio & 0.860 & 0.857 & 0.858 & 0.864 & 0.865 & 0.865 \\
    ROUGE Score & 0.864 & 0.867 & 0.869 & \textbf{0.868} & \textbf{0.868} & 0.867 \\
    \bottomrule
  \end{tabular}
  \end{adjustbox}
\end{table}

\subsection{Details for task 4: Clustering Performance}
\label{sec:task4}

\paragraph{Design Rationale:} Clustering evaluation provides a comprehensive assessment of semantic representation quality by testing whether similarity metrics preserve meaningful categorical structure in an unsupervised setting \citep{1427769}. V-measure provides a principled evaluation framework that harmonically combines cluster homogeneity (ensuring each cluster contains data points from a single class) and completeness (ensuring all data points from the same class are assigned to the same cluster) while maintaining invariance to cluster label permutations and symmetric properties under label exchange.

\paragraph{Datasets:} We utilize all 11 clustering datasets from the Massive Text Embedding Benchmark (MTEB) \cite{muennighoff2023mteb} (see Section \ref{sec:mteb_clustering} for complete details). These include: scientific literature clustering with ArXiv (S2S and P2P variants), BioRxiv (S2S and P2P), and MedRxiv (S2S and P2P, each with 17,647 rows) \cite{mteb_medrxiv_s2s_hf,mteb_medrxiv_p2p_hf}; community discussion clustering with StackExchange Standard (25 sets with 10-50 classes and 100-1000 sentences per class) and P2P (5 sets of 10k paragraphs, 5 sets of 5k paragraphs) \cite{mteb_stackex_standard_hf,mteb_stackex_p2p_hf}, Reddit Standard (199 subreddits across 25 sets) and P2P (10 sets of 50k, 40 sets of 10k paragraphs) \cite{mteb_reddit_standard_hf,mteb_reddit_p2p_hf}; and news clustering with TwentyNewsgroups (approximately 18,000 posts across 20 topics) \cite{mteb_twentyn_hf}. Documents range from 50 to 2,000 tokens with both balanced and imbalanced cluster distributions.

\paragraph{Evaluation:} We employ agglomerative hierarchical clustering with the following specifications:

\textit{Clustering parameters:}
\begin{itemize}
    \item Linkage criterion: Complete linkage (maximum distance between clusters)
    \item Distance computation: Metric-specific implementations
    \begin{itemize}
        \item Cosine: \(\text{distance} = 1 - (\text{embeddings} \cdot \text{embeddings}^T)\)
        \item Jaccard: \(\text{distance} = 1 - \frac{|\text{tokens}_i \cap \text{tokens}_j|}{|\text{tokens}_i \cup \text{tokens}_j|}\)
        \item ROUGE: \(\text{distance} = 1 - F_1\), where \(F_1 = \frac{2 \cdot \text{precision} \cdot \text{recall}}{\text{precision} + \text{recall}}\)
        \item BM25: TF-IDF based distance with parameters \(k_1=1.5\), \(b=0.75\), \(\epsilon=0.25\), normalized to [0,1]
        \item Others: \(\text{distance} = 1 - \text{similarity}\)
    \end{itemize}
    \item Number of clusters: Set to the ground-truth number of categories for each dataset
\end{itemize}

\medskip

\textit{V-measure computation:}

\[
    V = 2 \cdot \frac{\text{homogeneity} \cdot \text{completeness}}{\text{homogeneity} + \text{completeness}}
\]

where
\[
    \text{homogeneity} = 1 - \frac{H(C|K)}{H(C)}
    \quad \text{and} \quad
    \text{completeness} = 1 - \frac{H(K|C)}{H(K)},
\]
with \(H\) representing entropy, \(C\) the cluster assignments, and \(K\) the ground-truth classes. 

\medskip

We report V-measure scores ranging from 0 (random clustering) to 1 (perfect clustering alignment).

\subsection{Details for task 5: Retrieval Robustness}
\label{sec:task5}

\paragraph{Design Rationale:} Traditional retrieval evaluation assumes pristine textual conditions, yet real-world document corpora invariably contain OCR errors, typographical mistakes, formatting inconsistencies, and potentially malicious perturbations \cite{liu2024robustneuralinformationretrieval}. Our adversarial augmentation methodology comprehensively assesses retrieval robustness by evaluating similarity metrics' ability to maintain effectiveness when confronted with textual corruptions encountered in practical deployment environments \cite{wu2022pradapracticalblackboxadversarial}.

\paragraph{Datasets:} We utilize the complete BEIR benchmark \cite{thakur2021beir} (see Section \ref{sec:beir_benchmark}), comprising 18 standardized retrieval datasets across 9 IR task types. Key datasets include: MS MARCO with 8.84M documents and 6,980 queries, FEVER with approximately 5.42M Wikipedia passages, TREC-COVID with 171k documents and 50 queries, ArguAna with 8,674 documents and 1,406 queries, and CQADupStack with 457k documents and 13,145 queries \cite{beir_wiki,beir_paper_table}. Document lengths range from single sentences (20 tokens) to full articles (5,000+ tokens), providing comprehensive coverage of retrieval scenarios.

\paragraph{Evaluation:} We create adversarially augmented corpora through systematic perturbation:

\textit{Augmentation process:}
\begin{itemize}
    \item For each original document, generate 18 perturbed versions using transformations from Tasks 2 and 3.
    \item This increases corpus size by a factor of 19 (original + 18 perturbations).
    \item Apply transformations with reproducible implementation.
    \item Perturbations include: 
    \begin{itemize}
        \item Character-level and semantic alterations (6 types): sentence shuffling, word shuffling, negation toggling, character pruning (every 10th), random capitalization (25\%), and numerization
        \item Needle insertion (6 variations): 3 positions (0, 0.5, 1.0) $\times$ 2 sizes (15\%, 50\%)
        \item Content removal (6 variations): 3 positions (0, 0.5, 1.0) $\times$ 2 sizes (15\%, 50\%)
    \end{itemize}
\end{itemize}

\medskip

\textit{Performance measurement:}
\begin{itemize}
    \item Compute NDCG@10 (Normalized Discounted Cumulative Gain) for both original and augmented corpora.
    \item NDCG@10 calculation:
    \[
        \text{DCG@k} = \sum_{i=1}^{k} \frac{\text{rel}_i}{\log_2(i + 1)}
    \]
    where $\text{rel}_i$ is the relevance score of the document at rank $i$.
    \item Retention ratio:
    \[
        \text{Retention ratio} = 
        \frac{\text{NDCG@10}_{\text{perturbed}}}{\text{NDCG@10}_{\text{original}}}
    \]
    \item Aggregate using the harmonic mean across all perturbation types:
    \[
        H = \frac{n}{\sum_{i=1}^{n} \frac{1}{x_i}}
    \]
    where $x_i$ are the individual retention ratios.
    \item The harmonic mean is chosen to penalize poor performance on any single transformation more severely than the arithmetic mean.
\end{itemize}

\medskip

\textit{Implementation details:}
\begin{itemize}
    \item Embeddings are generated using cosine similarity for retrieval scoring.
    \item Queries with no relevant documents in the corpus are excluded from evaluation.
    \item The final robustness score is the harmonic mean of all retention ratios.
\end{itemize}

\subsection{Model and Metric Information}
\label{sec:models-metrics}

\subsubsection{Classical Text Similarity Metrics}

\paragraph{Levenshtein Ratio}
The implementation uses the \texttt{ratio} function from the \texttt{python-Levenshtein} library. For strings $x,y$ with edit distance $d_L(x,y)$ and lengths $|x|,|y|$:
\[
\operatorname{LevRatio}(x,y) = \frac{2 \cdot \text{matches}}{|x| + |y|} = \frac{(|x| + |y|) - d_L(x,y)}{|x| + |y|}
\]
where matches are the number of character matches in the optimal alignment.

\emph{Cost (per pair):} time $O(|x||y|)$; memory $O(\min\{|x|,|y|\})$.

\paragraph{ROUGE Score}
The implementation uses \texttt{rouge\_score} library with ROUGE-1 and ROUGE-2, returning the average of their F-measures:
\[
\text{ROUGE}_{\text{impl}} = \frac{\text{ROUGE-1}_{\text{fmeasure}} + \text{ROUGE-2}_{\text{fmeasure}}}{2}
\]
where ROUGE-1 measures unigram overlap and ROUGE-2 measures bigram overlap. The scorer uses \texttt{tiktoken} tokenizer (OpenAI's \texttt{text-embedding-3-small} tokenizer).

\emph{Cost:} tokenization + $n$-gram counting in time $O(\text{total tokens})$; memory $O(V_n)$ for $n$-gram maps.

\paragraph{Jaccard Similarity}
For text strings, the implementation first tokenizes using \texttt{tiktoken}, then computes Jaccard similarity on token sets $A,B$:
\[
J(A,B) = \frac{|A\cap B|}{|A\cup B|}, \qquad \text{distance} = 1-J(A,B)
\]

\emph{Cost (per pair):} $O(|A|+|B|)$ time after tokenization; memory $O(|A|+|B|)$.

\paragraph{BM25}
Uses \texttt{BM25Plus} from \texttt{rank\_bm25} library, a variant of BM25 with a delta parameter. For document $D$ and query $Q$:
\[
\mathrm{score}(D,Q) = \sum_{t\in Q}\mathrm{IDF}(t)\;
\frac{tf_{t,D}(k_1+1)}{tf_{t,D}+k_1\!\left(1-b+b\,\frac{|D|}{\mathrm{avgDL}}\right)} + \delta
\]
where $\delta$ is an additional parameter in BM25Plus. The implementation normalizes batch scores to $[0,1]$ range.

\emph{Cost:} similar to standard BM25; with inverted index, query-time proportional to postings traversed.

\paragraph{Cosine Similarity}
For embeddings $u,v\in\mathbb{R}^d$, implemented using PyTorch:
\[
\cos(u,v) = \frac{u\cdot v}{\|u\|_2\,\|v\|_2}
\]
\emph{Cost (per pair):} time $O(d)$; memory $O(d)$. With normalized embeddings where $\|u\|_2=\|v\|_2=1$, cosine equals dot product.

\subsubsection{Embedding Models (used with cosine similarity)}

\paragraph{Basic runtime \& storage notes.}
Embedding inference time scales roughly linearly with input tokens $L$; downstream similarity/retrieval scales with dimension $d$. Per-vector storage is $d\cdot b$ bytes (type size $b$: float32$=4$, float16$=2$, int8$=1$).

\paragraph{Models evaluated }
\begin{itemize}\setlength\itemsep{2pt}
  \item \textbf{OpenAI \texttt{text-embedding-3-small}}: default $d{=}1536$ (typical options: 512 or 1536); max input $\sim$8192 tokens. \emph{Cost:} pairwise cosine $O(d)$; storage $\propto d$.
  \item \textbf{OpenAI \texttt{text-embedding-3-large}}: default $d{=}3072$ (options: e.g., 256/1024/3072); max input $\sim$8192 tokens. \emph{Cost:} higher $d$ improves headroom at $\sim$2$\times$ storage vs 1536-D.
  \item \textbf{Cohere \texttt{embed-v4.0}}: $d\in\{256,512,1024,1536\}$ (default 1536); context up to 128k. Supports int8/uint8/binary outputs to reduce storage/I/O.
  \item \textbf{Voyage \texttt{voyage-3-large}}: default $d{=}1024$ (256/512/2048 options); context $\sim$32k. Offers compact dtypes (e.g., int8/binary).
  \item \textbf{Google \texttt{gemini-embedding-001}}: default $d{=}3072$ (typical 768/1536/3072 via \texttt{output\_dimensionality}); input limit $\sim$2048 tokens.
\end{itemize}

\subsection{NDCG Formula}
\label{sec:ndcg}

The Normalized Discounted Cumulative Gain at rank 10 (NDCG@10) is a metric that quantifies the ranking quality by giving higher importance to relevant documents at the top of the search results. The formula is defined as:
\[
\text{NDCG@10} = \frac{\text{DCG@10}}{\text{IDCG@10}}
= \frac{\sum_{i=1}^{10} \frac{2^{\text{rel}_i} - 1}{\log_2(i + 1)}}{\sum_{i=1}^{10} \frac{2^{\text{rel}^*_i} - 1}{\log_2(i + 1)}},
\]
where \(\text{rel}_i\) represents the relevance score of the item at position \(i\), and \(\text{rel}^*_i\) denotes the relevance score in the ideal ranking.

%%%%%%%%%%%%%%%%%%%%%%%%%%%%%%%%%%%%%%%%%%%%%%%%%%%%%%%%%%%%

\end{document}